%% file: main.tex
\documentclass[letterpaper, 10 pt, conference]{ieeeconf}  

\IEEEoverridecommandlockouts                              

\usepackage{amsmath} 
\usepackage{amssymb}  

\usepackage{caption}
\usepackage{subcaption}

\usepackage{verbatim}
\usepackage{xcolor}
\definecolor{dark-green}{RGB}{12,80,12}

\newcommand{\expnumber}[2]{{#1}\mathrm{e}{#2}}
\DeclareMathOperator{\E}{\mathbb{E}}
\usepackage{dsfont}

\usepackage{threeparttable}
\usepackage{xcolor}
\usepackage[export]{adjustbox}
\usepackage{array}
\usepackage{booktabs}
\usepackage{url}
\usepackage{siunitx}
\let\labelindent\relax
\usepackage{enumitem}
\setlist{nolistsep}

\usepackage{flushend}

\usepackage{tabularx}
\newcolumntype{Y}{>{\centering\arraybackslash}X}
\newcolumntype{Z}{>{\raggedleft\arraybackslash}X}

\newcommand{\para}[1]{\parskip=5pt \noindent\textit{#1:}}

\newcommand{\secref}[1]{Section~\ref{#1}}
\renewcommand{\eqref}[1]{Equation~(\ref{#1})}
\newcommand{\figref}[1]{Figure~\ref{#1}}
\newcommand{\tabref}[1]{Table~\ref{#1}}

\newcommand{\ours}{APFN}
\newcommand{\ourslong}{Active Particle Filter Networks}

\title{\LARGE \bf
Active Particle Filter Networks: Efficient Active Localization in Continuous Action Spaces and Large Maps
}

\author{Daniel Honerkamp, Suresh Guttikonda, and Abhinav Valada
\thanks{Department of Computer Science, University of Freiburg, Germany. This work was funded by the European Union’s Horizon 2020 research and innovation program under grant agreement No 871449-OpenDR.}
}

\begin{document}

\maketitle
\thispagestyle{empty}
\pagestyle{empty}

\begin{abstract}

Accurate localization is a critical requirement for most robotic tasks. The main body of existing work is focused on passive localization in which the motions of the robot are assumed given, abstracting from their influence on sampling informative observations. While recent work has shown the benefits of learning motions to disambiguate the robot's poses, these methods are restricted to granular discrete actions and directly depend on the size of the global map. We propose \ourslong{} (\ours), an approach that only relies on local information for both the likelihood evaluation as well as the decision making. To do so, we couple differentiable particle filters with a reinforcement learning agent that attends to the most relevant parts of the map. The resulting approach inherits the computational benefits of particle filters and can directly act in continuous action spaces while remaining fully differentiable and thereby end-to-end optimizable as well as agnostic to the input modality. We demonstrate the benefits of our approach with extensive experiments in photorealistic indoor environments built from real-world 3D scanned apartments. Videos and code are available at \url{http://apfn.cs.uni-freiburg.de}.

\end{abstract}

\section{Introduction}
\input{1_intro}

\section{Related Work}
\input{2_related_work}

\section{\ourslong{}}
\input{3_approach}

\section{Experimental Results}
\input{4_experiments}

\section{Conclusion}
\input{5_conclusion}
\bibliographystyle{IEEEtran}
\bibliography{IEEEabrv.bib,references.bib}
\end{document}

%% file: 1_intro.tex
The ability of a robot to accurately localize itself is a core requirement across almost all robotic tasks from navigation~\cite{hurtado2021learning, younes2021catch} to mobile manipulation~\cite{desouza2002vision, honerkamp2022n, honerkamp2021learning}. Accordingly, a broad body of research has been devoted to this topic. The by far most common approach is to first define an initial guess of the robot's pose, then manually move the robot until the localization algorithm has roughly converged and continue to constantly localize the robot while it executes its tasks. This is known as passive, local localization.

Most localization algorithms rely on a form of feature matching between the current observations and a given (2D) map of the environment. As such their performance strongly depends on the current observations, which in turn are decided by the robot's motions which decide what parts of the map will be observed. But the ability to sample informative observations has remained largely unexplored. In this work, we investigate the benefits of \textit{active localization}, in which the robot can actively seek observations that are most informative of its current pose in the environment. Furthermore, the agent can counteract the strengths and weaknesses of particular localization modules by actively avoiding ambiguous situations and failure modes of the localization module.

\input{figures/tex/teaser}



Previous work has extended Adaptive Markov Localization to active control by greedily maximizing information theoretic quantities~\cite{fox1998active, fox1999markov}, but for the most part, remained restricted to analytical observation models and structured observations. More recently, learning-based methods have shown the benefits of active decision making for localization~\cite{chaplot2018active, gottipati2019deep}, though have remained constrained to simple environments~\cite{chaplot2018active} or discrete actions and small maps~\cite{gottipati2019deep}, having to process the global map at every possible orientation of the agent at each step. 

We present an approach that couples probabilistic and learning-based methods through learned particle filters~\cite{karkus2018particle} and deep reinforcement learning (RL) to generalize to continuous action spaces and arbitrary sensor modalities independent of map size. Particle filters \cite{thrun2001robust} enable efficient representation of multi-modal beliefs over large maps. These mechanisms can be made fully differentiable~\cite{karkus2018particle, Jonschkowski2018differentiable}, enabling us to learn the components of a particle filter end-to-end, thereby extending it to abstract observations such as pixels or depth maps. Importantly, these networks only need to process local information for each particle. We then train a reinforcement learning agent that selects actions to minimize the overall localization error, following the same principle of processing only local information over the most likely hypotheses through a hard attention mechanism. In contrast to previous work, this enables us to process hypotheses over continuous poses~\cite{chaplot2018active, gottipati2019deep} while at the same time breaking the dependency on processing the full map with a neural network.

We evaluate our approach in extensive photorealistic scenes of real-world 3D scanned apartments from the gibson dataset~\cite{xia2018gibson} in the iGibson simulator~\cite{chengshu2021igibson2} and find substantial improvements in localization error over the baselines, demonstrating the benefits of the learned policy.

To summarize, this work makes the following main contributions:
\begin{itemize}
    \item We leverage the combination of probabilistic principles with learned methods to achieve a very flexible and fully differentiable approach for active localization which does not depend on a specific sensor modality.
    \item We break the dependency of learning-based approaches on action granularity and map size, enabling the approach to work in continuous action spaces and arbitrary map sizes.
    \item We demonstrate the benefits of this approach in large photo-realistic indoor environments built from real-world 3D scanned apartments.
    \item We make the code publicly available at \url{http://apfn.cs.uni-freiburg.de}.
\end{itemize}


%% file: figures/tex/teaser.tex
\begin{figure}
    \footnotesize
	\centering
	\resizebox{\columnwidth}{!}{%
  		\includegraphics[width=0.32\columnwidth,trim={0.0cm 0.0cm 0.0cm 0.0cm},clip,angle =0]{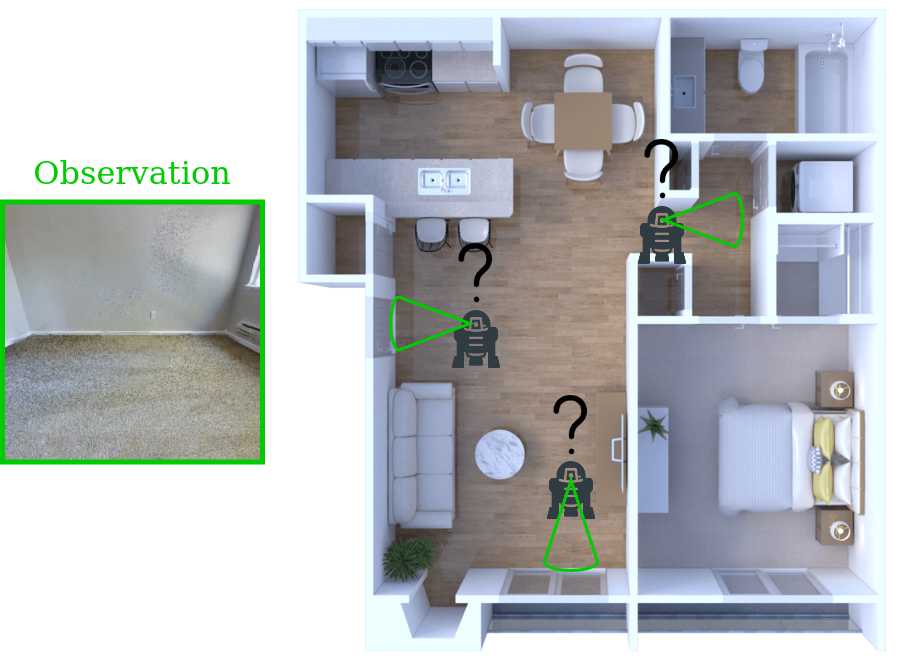}
	}
	\caption{To localize itself, a robot has to match its observations with a given map of the environment. To distinguish ambiguous poses requires to find observations that maximally disambiguate among the true pose given the robot's belief over its current pose. To do so, certain trajectories through the apartment reveal clearly more information than others. We propose \ourslong{} which combines learned particle filters with active decision making to sample the most informative observations.}
  	\label{fig:teaser}
\end{figure}

%% file: 2_related_work.tex
Localization is a well studied field with a long-standing history. In the following, we discuss both passive and active localization methods which have been tackled using classical and learning-based techniques.

\para{Passive Localization} 
A large number of established localization approaches rely on Bayesian filtering-based techniques. These include methods based on Kalman filters~\cite{smith1990estimating} which are restricted to modeling unimodal (Gaussian) beliefs, Multi-Hypothesis Kalman filters that use mixtures of Gaussians~\cite{cox1994modeling} and non-parametric particle filters which can model arbitrary distributions.
Particle filters are widely used in methods such as Monte Carlo Localization and Adaptive Monte Carlo Localization (AMCL)~\cite{thrun2001robust}. Though these methods usually rely on structured observations and analytic observation models and therefore are most commonly used with LiDAR observations. While there are approaches that incorporate depth or camera images~\cite{dellaert1999using, coltin2013multi}, constructing observation models for them is extremely challenging.
Recently, fully differentiable versions of particle filters have been introduced~\cite{karkus2018particle, Jonschkowski2018differentiable}. These fully differentiable versions enable the use of arbitrary modalities through end-to-end optimization. LASER extends MCL with learned circular features and rendering in latent space~\cite{min2022laser}. Learning-based methods have also been proposed to extract explicit features such as room layout edges~\cite{boniardi2019robot} or to estimate odometry directly from visual inputs~\cite{clark2017vinet, vodisch2022continual}.

\para{Active Localization} 
Active localization has received comparably little attention in the past. Active versions of both Markov Localization~\cite{fox1998active, fox1999markov} and Kalman filters~\cite{jensfelt2001active} have been proposed. These methods inherit the need for structured observations or expert-specified observation models and as such cannot easily incorporate contextual clues or high-dimensional observations. Their objectives are to maximize information theoretic quantities such as the reduction in entropy of the belief.
Chaplot~\textit{et~al.}~\cite{chaplot2018active} introduce a learnable Bayesian filtering approach in combination with reinforcement learning. While they are able to learn good active policies, the model relies on access to observations from across the environment to compute features ahead of time and at each step has to process the full map for every possible discrete orientation. As a consequence, the approach does not easily generalize to different map sizes at test time and does not scale well to large maps or continuous actions. Gottipati~\textit{et~al.}~\cite{gottipati2019deep} introduce a hierarchical likelihood model in which the full map only has to be processed at a coarse resolution and only likely areas are processed at higher resolutions. Nonetheless, the dependency on the map size remains and only discrete actions can be evaluated. For both approaches, the dimensionality of the reinforcement learning agent's inputs scales linearly with the discretization of the rotation actions.
In contrast, our approach never has to process the full map with a neural network and can directly evaluate continuous poses and actions.

\para{Active SLAM} In simultaneous localization and mapping both information gain objectives~\cite{feder1999adaptive, leung2006active} and reinforcement learning based approaches~\cite{chaplot2020learning, placed2020deep} have been used to control a robot while performing both the mapping and localization. In contrast to active localization, these approaches do not have access to the prior knowledge of the map to steer the robot towards informative states~\cite{placed2022survey}.


%% file: 3_approach.tex
We propose \ourslong{} (\ours{}) consisting of two modules: a learned particle filter network maintaining a distribution over the current belief of the robot's pose and a reinforcement learning agent taking decisions based on the current observations and belief. An overview of our approach is depicted in \figref{fig:scheme}.

\input{figures/tex/scheme}

In the following, we will define the active localization task that we aim to solve and then introduce our approach.

\subsection{Problem Statement}

We assume a mobile robot that receives exteroceptive sensor readings $sens_t$ and proprioceptive odometry measurements $m_t$, placed randomly in an environment. Given a map $M$ of the environment, we seek the sequence of actions $a_{1:T}$ that minimizes the pose error of the robot over a fixed time horizon $T$.
We may be given an initial guess of the initial pose of the robot (local localization) or have to start from a uniform belief over the full map (global localization).

\subsection{Localization Module}
The robot starts with an initial belief $b_0$, either uniformly distributed over the map or based on an initial guess.
Given the current observation $o_t = [sens_t, m_t]$, we then use a differentiable particle filter network (PF-net)~\cite{karkus2018particle} to update the current belief over the robot's pose. PF-net uses neural networks to present the observation and transition model of a particle filter. By using a soft-resampling, where new particle weights $w^{'k}_t$ of $K$ particles are sampled from a distribution 
\begin{equation}
    q(k) = \alpha w^k_t + \frac{(1 - \alpha) }{K}
\end{equation}
the gradients are non-zero for values of $\alpha~\neq~1$, enabling us to optimize through the whole network.
The observation model calculates the likelihood $f_t^k$ of a particle based on an encoding of the current sensor readings and particle-centric local map which is extracted from the global map through a differentiable spatial transformer module \cite{jaderberg2015spatial}. As a result, the likelihood of each particle can be evaluated based on local information without the need to process the full global map.

This provides a number of advantages for active localization: 
(i) the network is fully differentiable and thereby can be jointly optimized with deep reinforcement learning algorithms,
(ii) it is flexible to arbitrary robot sensors, making it applicable to a wide range of robotic platforms and
(iii) it can handle continuous actions and arbitrary map sizes.

The model is trained end-to-end to minimize the mean squared pose error
\begin{equation}
    \mathcal{L}_{pfnet} = \sum_t (\hat{x}_t - x_t^*)^2 + (\hat{y}_t - y_t^*)^2 + \beta(\hat{\phi}_t - \phi_t^*)^2,
\end{equation}
where $\hat{x}, \hat{y}, \hat{\phi}$ and $x^*, y^*, \phi^*$ are estimated and ground-truth pose of the robot and $\beta$ is weighting term. We follow the architecture of the original work~\cite{karkus2018particle} which uses convolutional encoders for both the raw observations and the local maps, then process the concatenated features with a stack of locally fully-connected and fully-connected layers.

\subsection{Active Localization}

We aim to learn a policy to move the agent such that, given the current belief about the robot's pose and the localization module, it can best disambiguate the true pose.
The agent is operating in a Partially Observable Markov Decision Process~(POMDP) $\mathcal{M} = (\mathcal{S}, \mathcal{A}, \mathcal{O}, \mathcal{T}(s' | s, a), P(o | s), r(s, a))$ where $\mathcal{S}, \mathcal{O}$ and $\mathcal{A}$ are the state, observation and action spaces, $\mathcal{T}$ and $P$ describe the transition and observation probabilities, and $r$ and $\gamma$ are the reward and discount factor. The agent's objective is to learn a policy $\pi(a | \cdot)$ that maximises the expected return $\E_\pi[\sum_{t=1}^{T} \gamma^t r(s_t, a_t)]$.

\para{Belief Representation}
As the ground truth robot pose is not directly observable, the agent has to act based on its current belief over the state.
The PF-net provides us with a multi-modal belief $b_t$ over the global map, represented by the particle state. We transform this into a spatial, permutation invariant representation by projecting the particles into a belief map of dimension $H \times W \times 4$ where $H$ and $W$ are the height and width of the global map and the first channel is the occupancy map, the second channel is the aggregated weights for all particles in a given cell and the third and fourth channel are the weighted sine and cosine of all particles in a given cell. The sine and cosine are used to circumvent the non-linearity in the angles.

\para{Agent}
We propose a reinforcement learning agent that observes both its current belief together with the low-level robot observations and learns a policy $\pi(a_t | b_t, o_t; l)$ where $l$ is the localization module. This allows it to improve the localization in two ways: (i) actively sample the most informative sensor readings and (ii) take into account the localization module's strengths and weaknesses, e.g. avoid observations where the localization module does not perform well.\looseness=-1

While the belief stretches the full map, we find that within very few steps the particles concentrate on a small number of most likely regions. As such we apply the principle of local information to break the dependency on the full map. To do so, we extract local maps around the modes of the particle distribution from the belief map. The agent then observes a stack of $k$ local belief maps, each centered and oriented according to a mode of the distribution. This is akin to a hard attention mechanism, which can be made fully differentiable if desired \cite{xu2015show}. In practice, we find that just using the mean position and orientation of the particles works well, but extending this to cover the top $k$ modes is straightforward. In contrast to previous work, this allows us to process and generalize to arbitrary map sizes and arbitrary continuous poses and actions.

The agent is trained to directly minimize the prediction error of the localization network. At each step, it receives a reward
\begin{equation}
    r = - \mathcal{L}_{pfnet} - \lambda_{collision} * \mathds{1}_{collision},
\end{equation}
where $\mathcal{L}_{pfnet}$ is the prediction loss of the PF-net, $\mathds{1}_{collision}$ is a binary collision indicator and $\lambda_{collision}$ is a weighting constant. The agent has a fixed number of environment steps to localize itself, after which the episode terminates.

\para{Training} 
While the approach is fully differentiable and can be optimized end-to-end, we find it beneficial to pretrain the localization network for better stability. Though joint finetuning may be able to further improve results. For pretraining we use a goal-reaching agent (see \secref{sec:exp_setup} for details) to collect a dataset of 4,000 episodes of length 25 and then perform supervised training following Karkus~\textit{et~al.}~\cite{karkus2018particle}, using a tracking task with only 30 particles.
We train the RL agent with soft-actor critic (SAC) \cite{haarnoja18b}, which has been shown to produce strong policies in continuous control and robotics tasks. Hyperparameters for all components are reported in \tabref{tab:hyper}.\looseness=-1
\input{tables_new/hyperparameters}



%% file: figures/tex/scheme.tex
\begin{figure*}
	\centering
	\resizebox{.7\textwidth}{!}{%
  		\includegraphics[width=\columnwidth,trim={0.0cm 0.0cm 0.0cm 0.0cm},clip,angle =0]{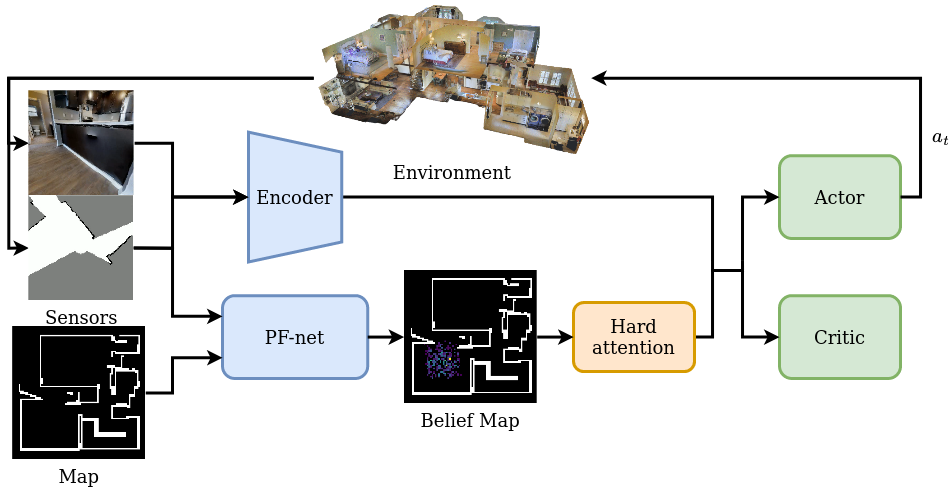}
	}
	\caption{Overview of the proposed \ourslong{}. Given the robot's observations, the PF-net updates the belief over the current pose of the robot, modeled as a particle distribution. This distribution is then projected into a belief map over the environment. The RL agent attends to the local regions across the most likely hypotheses as well as the raw robot observations and produces actions $a_t$ to move the robot's base, which then result in the next sensory observations for the PF-net.}
  	\label{fig:scheme}
\end{figure*}

%% file: tables_new/hyperparameters.tex
\begin{table}
    \centering
    \begin{tabularx}{0.485\textwidth}{l|YY}
      \toprule
        Parameter & PF-net & RL \\
        \midrule
        Train steps & 400,000 & 1,000,000   \\
        batch size       & 8 & 256\\
        lr               & $\expnumber{2.5}{-3}$ & $\expnumber{3}{-4}$\\
        resample         & false & true \\
        $\alpha$         & -- & 0.5 \\
        $\beta$          & 0.36 & 0.36\\
        $T$              & 25 & 50 \\
        particles        & 30 & 500 \\
        initial distribution          & tracking & semi-global \\
        initial std translation       & 0.3 & 0.3 \\
        initial std angular           & $\pi / 6$ & $\pi / 6$ \\
        transition noise translation  & 0.0 & 0.01 \\
        transition noise angular      & 0.0 & $\pi / 36$ \\
        control frequency             & 1.7Hz & 1.7Hz \\
        $\tau$           & -- & 0.005 \\
        $\gamma$            & -- & 0.99\\
        replay buffer size     & -- & 50,000 \\
        entropy coefficient     & -- & learned \\
        $\lambda_{collision}$  & -- & 0.1 \\
      \bottomrule
    \end{tabularx}
    \caption{Training hyperparameters for the PF-net (left) and the reinforcement learning agent (right).}
    \label{tab:hyper}
    \vspace{-0.2cm}
\end{table}

%% file: 4_experiments.tex
To evaluate the effectiveness of our approach, we perform extensive experiments in photorealistic indoor environment. With these experiments we aim to answer the following
questions:
\begin{itemize}
    \item Does the PF-net localization module scale to complex, photorealistic indoor environments and generalize to unseen apartments in these settings?
    \item Does our proposed approach learn to localize itself in both seen and unseen apartments and across different tasks from local to global localization?
    \item Is the learned policy able to find trajectories that achieve better localization than alternative control policies?
\end{itemize}

\subsection{Experiment Setup}\label{sec:exp_setup}
To evaluate our approach, we train a LoCoBot robot in the photorealistic iGibson simulator~\cite{chengshu2021igibson2}. The LoCoBot robot has a differential drive and is equipped with an RGB-D camera with a field-of-view of \SI{90}{\degree} and a maximum depth of \SI{10}{\meter} as well as a LiDAR with a range of \SI{240}{\degree}. The action space consists of the linear and angular velocities for the base. We use a subset of 45 apartment scenes from the gibson dataset~\cite{xia2018gibson}, split into 38 training and 7 unseen test apartments. The test apartments are completely unseen by both the PF-net and the RL agent.


\para{Baselines}
To evaluate the policy of the reinforcement learning agent, we compare our approach against the following baselines:
\begin{itemize}
    \item \textit{Avoid:} A simple heuristic policy that drives forward until its depth camera recognizes a close object. We divide the depth image into four horizontal quarters and, depending on in which quarter of the depth image the close object is, drive backwards or turn away from the obstacle.
    \item \textit{Goalnav:} A policy that navigates towards a random target in the environment. It uses a path-planner based on access to the ground truth traversability map and robot pose to reach this goal.
    \item \textit{Turn:} An agent that always turns in place at maximum angular velocity.
\end{itemize}

\para{Tasks}
We focus on three localization tasks, ranging from local to global localization. These are
\begin{itemize}
\item \textit{Tracking:} the initial particles are sampled from a multi-variate Gaussian distribution with a standard deviation of \SI{0.3}{\meter} and \SI{30}{\degree} and centered at a random pose sampled with the same standard deviations around the ground truth robot pose. The PF-net uses 300 particles.
\item \textit{Semi-global localization:} We uniformly sample 500 particles from a box of 3.3$\times$\SI{3.3}{\meter} around the initial guess.
\item \textit{Global localization:} We sample 3,000 particles uniformly across the traversable area of the whole map.
\end{itemize}

\para{Metrics} We report the root mean squared positional error in centimeters and root mean squared angular error in radians, referred to as \textit{position} and \textit{orient} in the tables. All metrics are averaged over 50 episodes.

\subsection{Passive Localization}
\input{tables_new/passive}

\input{tables_new/active}
\input{figures/tex/episode}

The original PF-net model has focused on evaluation in the simpler, static House3D dataset \cite{wu2018building}. We implement a version of this model based on the author's code for the iGibson simulator and evaluate it on scenes from the photorealistic gibson dataset, which are based on real-world 3D scans of apartments. We report the results for passive localization for different modalities based on the goalnav agent that collected the training data in \tabref{tab:passive}. LiDAR scans are converted to occupancy maps in which 0 is free space, 1 unexplored, and 2 occupied.

We find that the PF-net performs well in these more complex scenes, achieving a positional error of around \SI{20}{}-\SI{25}{\centi\meter} for tracking, which, for both modalities, is actually lower than the \SI{40}-\SI{49}{\centi\meter} error reported on the House3D dataset \cite{karkus2018particle}. 
Moreover, the network generalizes well to unseen apartments, showing no significant generalization gap.
Even though the field-of-view of the RGB-D camera is much smaller than what the LiDAR can sense, both modalities achieve relatively similar performance, highlighting that the network is able to extract rich information in the complex pixel observations.

\subsection{Active Localization}

For active localization, we focus on the best performing LiDAR modality. The RL agent observes the robot state consisting of current forward and angular velocities, a collision flag, and the remaining steps in the episode together with the occupancy map from the LiDAR and the local belief map. To ensure it can learn full obstacle avoidance, it also receives the RGB-D observations. The policy consists of a shared feature encoder, made up of three convolutional networks, one for RGB-D and one for the occupancy map. Each network consists of layers with (channels, kernel size, stride) of [(32, (3, 3), 2), (64, (3, 3), 2), (64, (3, 3), 1), (64, (2, 2), 1)].
These features are then concatenated with the robot state and passed to the actor and critic, consisting of a two-layer MLP with 512 neurons. All intermediate layers are followed by ReLU activations. All pixel-based observations are of size $56 \times 56$. While we train the PF-net on ground truth odometry data, during the policy training we add zero-centered Gaussian noise with standard deviation of \SI{1}{\centi\meter} and \SI{5}{\degree} to the transitions. We train the policy with 500 particles and at test time evaluate with varying numbers of particles as defined for the different tasks.

\tabref{tab:new_results_final} reports the results for the active localization tasks for both seen and unseen apartments. First of all, we find large differences in localization performance across the different motion models. This highlights the strong dependence on the robot's movements and confirms the importance of active decision making for globalization. We find that the reinforcement learning agent consistently achieves the best localization across all tasks. The only exception is the positional error in the tracking task, in which the turn policy achieves a very low positional error, but suffers from a large angular error. Note that in this task the initial particle distribution is already fairly accurate, as such it may actually be beneficial to remain in place.
Moreover, the agent successfully generalizes to unseen apartments. Note that these apartments have not been seen by both the RL agent and the localization module. To succeed in these apartments, the agent has to learn general movement patterns and the ability to seek out informative regions. 
Lastly, we find that differences in localization performance are particularly large in the global localization task with our approach reducing the positional error by over 60\% in comparison to other motions. This is expected as in global localization we have the least amount of prior information about the robot's pose.

Qualitatively we find that the reinforcement agent performs targeted movements through the room with frequent rotations which reveals a large area of the apartments and is aligned with the strong performance of the turn baseline. Examples of the agent's trajectories and inputs are shown in \figref{fig:episode} and in the accompanying video.


%% file: tables_new/passive.tex
\setlength{\tabcolsep}{2pt}
\begin{table*}
    \centering
    \begin{tabularx}{\textwidth}{l|YY|YY|YY||YY|YY|YY}
      \toprule
        Task & \multicolumn{6}{c||}{seen} & \multicolumn{6}{c}{unseen}\\
        \cmidrule{2-13}
        & \multicolumn{2}{c}{Tracking} & \multicolumn{2}{c}{Semi-Global} & \multicolumn{2}{c||}{Global} & \multicolumn{2}{c}{Tracking} & \multicolumn{2}{c}{Semi-Global} & \multicolumn{2}{c}{Global}\\
        \cmidrule{2-13}
        Modality     & position & orient & position & orient & position & orient & position & orient & position & orient & position & orient \\
      \midrule
         LiDAR & 20.8 & 0.13 & 27.2 & 0.21 & 111.4 & 0.33 & 18.9 & 0.12 & 23.7 & 0.16 & 141.2 & 0.38\\
         RGB-D & 24.8 & 0.15 & 30.2 & 0.20 & 126.6 & 0.30 & 24.5 & 0.16 & 29.2 & 0.18 & 144.1 & 0.34\\
      \bottomrule
    \end{tabularx}
    \caption{Passive localization results on the iGibson dataset for different localization tasks. We report the average root mean squared positional error in centimeter (\textit{position}) and the root mean squared orientation error of the robot's yaw in radians (\textit{orient}). Evaluated with the pretraining settings for $T=25$ timesteps.}
    \label{tab:passive}
\end{table*}
\setlength{\tabcolsep}{6pt}

%% file: tables_new/active.tex
\setlength{\tabcolsep}{2pt}
\begin{table*}
    \centering
    \begin{tabularx}{\textwidth}{l|YY|YY|YY||YY|YY|YY}
      \toprule
        Task & \multicolumn{6}{c||}{seen} & \multicolumn{6}{c}{unseen}\\
        \cmidrule{2-13}
        & \multicolumn{2}{c}{Tracking} & \multicolumn{2}{c}{Semi-Global} & \multicolumn{2}{c||}{Global} & \multicolumn{2}{c}{Tracking} & \multicolumn{2}{c}{Semi-Global} & \multicolumn{2}{c}{Global}\\
        \cmidrule{2-13}
        Agent & position & orient & position & orient & position & orient & position & orient & position & orient & position & orient \\
      \midrule
         Goalnav     &  16.8 & 0.12 & 18.2 & 0.12 & \phantom{0}99.3 & 0.24 & 14.9 & 0.11 & 21.4 & 0.15 & 113.3 & 0.21 \\
         Avoid       &  15.8 & 0.13 & 22.4 & 0.15 & 152.0 & 0.32 & 15.8 & 0.12 & 33.5 & 0.19 & 162.9 & 0.29 \\
         Turn        &  \textbf{11.8} & 0.80 & 14.6 & 0.09 & 103.1 & 0.30 & 13.9 & 0.10 & 19.8 & 0.12 & 115.9 & 0.31 \\
         \ours{} (ours) &  13.4 & \textbf{0.10} & \textbf{11.7} & \textbf{0.08} & \textbf{\phantom{0}74.8} & \textbf{0.16} & \textbf{11.1} & \textbf{0.08} & \textbf{16.3} & \textbf{0.11} & \textbf{\phantom{0}63.3} & \textbf{0.17} \\
      \bottomrule
    \end{tabularx}
    \caption{Active localization results in the iGibson simulator in seen and unseen apartments. The localization module is based on LiDAR occupancy maps. We report the average root mean squared positional error in centimeter (\textit{position}) and the root mean squared orientation error of the robot's yaw in radians (\textit{orient}).}
    \label{tab:new_results_final}
\end{table*}
\setlength{\tabcolsep}{6pt}

%% file: figures/tex/episode.tex
\begin{figure*}
	\centering
	\footnotesize
	{\setlength{\fboxsep}{0pt}%
  \setlength{\fboxrule}{0pt}%
	\resizebox{\textwidth}{!}{%
  	\begin{tabular}{c}
  		\fbox{\includegraphics[width=\textwidth,trim={0.0cm 0.0cm 0.0cm 0.0cm},clip,angle =0]{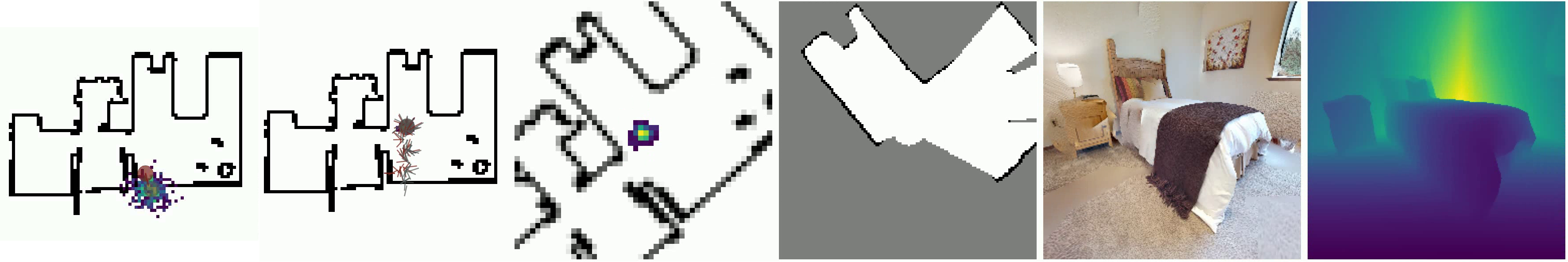}}\\
  		\fbox{\includegraphics[width=\textwidth,trim={0.0cm 0.0cm 0.0cm 0.0cm},clip,angle =0]{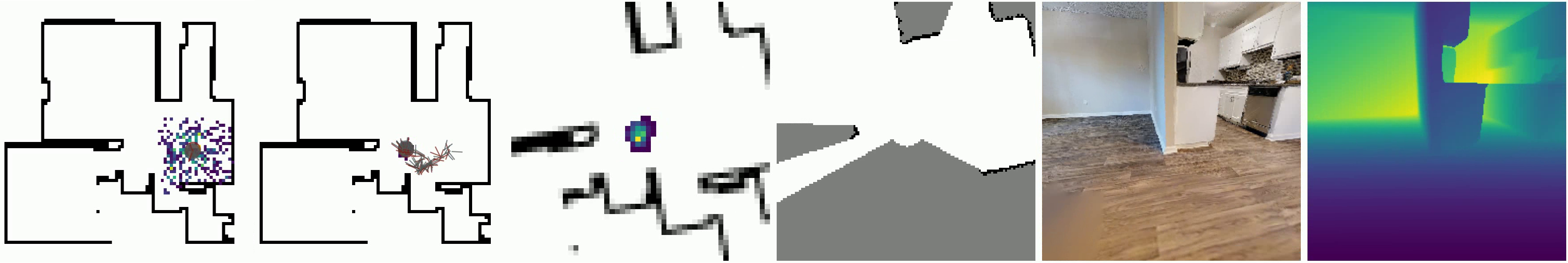}}\\
  		\fbox{\includegraphics[width=\textwidth,trim={0.0cm 0.0cm 0.0cm 0.0cm},clip,angle =0]{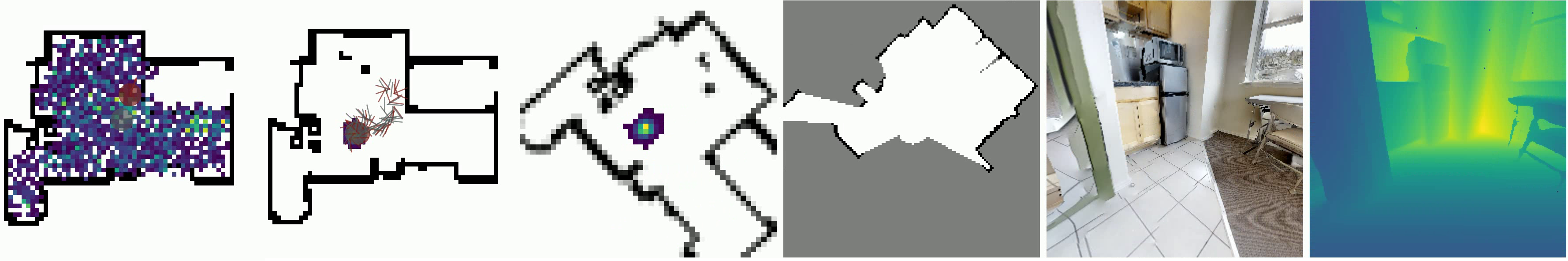}}
  	\end{tabular}
	}}
	\caption{Examples for the tracking (top), semi-global (mid) and global (bottom) localization tasks. Left: the initial particle distribution, second from left: the global map and trajectory of the agent. Green arrows denote the estimated poses and red the ground truth poses at each step. Circles denote the final estimated and ground-truth poses. Third from left to right: the local belief map observed by the RL agent, the current observations: occupancy grid, RGB and depth.}
  	\label{fig:episode}
\end{figure*}

%% file: 5_conclusion.tex
In this work, we introduce \ourslong{} which combines probabilistic filtering methods with learned decision making to accurately localize a robot in realistic indoor environments. In contrast to previous methods, our approach scales to continuous action spaces and arbitrary map sizes by selectively attending to only local information. In extensive experiments, we evaluate this ability in photorealistic indoor environments and find that it is able to accurately localize itself in both seen and completely unseen apartments. The learned policy considerably outperforms the baselines, demonstrating strong improvements in localization performance by sampling informative observations.

In future work, we aim to extend the approach to simultaneously control sensors such as actuated cameras, which promises to benefit even more from active perception. Another promising avenue is the extension of learning-based localization and attention mechanisms to dynamic environments and noisy, partial or incorrect maps in which it becomes important to selectively filter out uncertain or incorrect observations. Lastly, the trade-off between active localization and other task objectives is an exciting direction for future work.